%% file: EgoHandWAM.tex
\RequirePackage[T1]{fontenc}
\documentclass[journal,10pt]{IEEEtran}

\usepackage{amsmath,amssymb,booktabs,array,graphicx,xcolor,calc}
\usepackage{tikz}
\renewcommand{\IEEEkeywordsname}{Keywords}
\usepackage[hidelinks]{hyperref}
\IfFileExists{xurl.sty}{\usepackage{xurl}}{}
\input{header.tex}
\title{EgoHandWAM: Probing the Role of Hand Motion in Egocentric World\textendash{}Action Modeling}
\input{authors.tex}
\date{}
\hypersetup{pdftitle={EgoHandWAM: Probing the Role of Hand Motion in Egocentric World\textendash{}Action Modeling},pdfauthor={Kailin Wang, Haoxiang Jie, Yaoyuan Yan, Zhiyou Heng, Zhaosong Li}}
\begin{document}
\maketitle
\begin{abstract}
How much does hand motion tell us about what a first-person camera will see next? EgoHandWAM forecasts three-dimensional wrist and fingertip trajectories and a future visual feature from the current image and recent hand states. We train on 751,065 EgoDex windows using a frozen DINOv2 encoder. On held-out development sessions, the action-only model reduces two-second wrist endpoint error from 60.26 to 56.48~mm relative to a stationary forecast. It still underestimates long reaches. Adding future-feature supervision gives nearly identical trajectory accuracy. With the recorded hand history, the visual head reduces smooth L1 error by 1.32\% relative to copying the current feature. Shuffling histories across sessions erases this gain. In a separate four-task probe, adding recorded future displacements to observed context and future tracking-validity masks reduces visual-feature error by 3.93\%. Masks alone give little improvement. A trajectory predicted from the same observations adds little. On two simulated RoboTwin tasks, substituting aligned human examples for an auxiliary batch of robot examples gives no consistent increase in task success across three training seeds.

\end{abstract}
\section{Introduction}\label{introduction}

When someone reaches for a lid, a first-person camera records the hand, the object and the motion of the camera itself. A wrist trajectory captures part of this interaction. How much can a model anticipate before the hand moves, and how useful is that information for predicting what comes into view? Ego4D and EgoDex make it possible to study these questions in recorded human activity \cite{ref-ego4d,ref-egodex}.

We examine hand motion in three roles. First, motion is a \emph{prediction target}. Given the current image and eight observed hand states, EgoHandWAM predicts the next two seconds of wrist and fingertip positions. A second head predicts the visual feature at the last time step. We vary the weight of the visual loss while holding the trajectories, inputs and training budget fixed.

Second, motion is an \emph{input to a visual forecast}. A separate regression model receives either a recorded future trajectory with its tracking mask or a trajectory predicted from the observed context. Comparing these inputs distinguishes information available after the movement from information a model can supply at prediction time.

Third, human recordings provide \emph{auxiliary training data for a robot policy}. A shared Transformer learns from human and robot observations, while simulated demonstrations supervise the robot commands. We measure task completion in simulated bottle grasping and bimanual block handover, and test whether the correspondence between a human hand history and its video matters. Only robot demonstrations supply executable action labels. Together, these experiments trace how far a hand signal that helps forecast a visual feature carries over to trajectory accuracy and robot task success.

\section{Related Work}\label{related-work}

\subsection{Egocentric data and hand
forecasting}\label{egocentric-data-and-hand-forecasting}

Ego4D establishes a broad setting for understanding and forecasting
first-person activity \cite{ref-ego4d}. EgoDex provides manipulation video with
3D hand tracking and an official two-second forecasting comparison for the same wrist and fingertip landmarks \cite{ref-egodex}. Its benchmark includes task language and a fixed test split. We use session-held-out development data to compare supervision conditions. Some recordings lack confidence fields \cite{ref-egodexformat}.

EPIC-KITCHENS-100 expands fine-grained action recognition and
anticipation in everyday environments \cite{ref-epic100}, while Ego-Exo4D links
first- and third-person observations of skilled activity \cite{ref-egoexo4d}.
Geometric datasets provide complementary supervision: H2O annotates both
hands and manipulated objects \cite{ref-h2o}; AssemblyHands supplies
multi-view-derived hand pose annotations \cite{ref-assemblyhands}; HOT3D supports
egocentric hand-object tracking \cite{ref-hot3d}; and ARCTIC captures bimanual
manipulation of articulated objects \cite{ref-arctic}. These resources differ in
capture setting, geometry and task definition. We use EgoDex for the human forecasting experiments and specify how we select windows with usable tracking labels.

Hand pose recovery and future trajectory prediction also have different
input-output requirements. WildHands studies pose estimation in everyday
egocentric images, including the effects of perspective and camera
information \cite{ref-wildhands}. HaMeR reconstructs hand meshes from monocular
images with a transformer-based model \cite{ref-hamer}. The Object-Centric Transformer
jointly predicts hand motion and interaction hotspots from egocentric
video \cite{ref-oct}, connecting future motion with likely interaction
locations. Our target is instead a fixed set of future three-dimensional
wrist and fingertip positions.

Prior trajectory forecasting methods model temporal dependencies and
uncertainty. The uncertainty-aware state space transformer studies
three-dimensional hand forecasting from observed egocentric RGB video
\cite{ref-usst}. EgoMAN considers three-dimensional hand trajectory prediction
from human interaction video \cite{ref-egoman}. EgoH4 distinguishes global wrist
motion from articulated hand prediction and evaluates motion baselines
\cite{ref-egoh4}. These works motivate geometric displacement and position
errors. Our observed hand-state inputs, landmark set, horizon and
coordinate protocol differ, so their published scores are not treated as
directly comparable baselines.

\subsection{Visual representations for world-action
learning}\label{visual-representations-for-world-action-learning}

Visual pretraining objectives offer different forms of information. DINO
learns by self-distillation without labels \cite{ref-dino}, masked autoencoders
reconstruct missing image patches \cite{ref-mae}, and CLIP aligns images with
natural-language descriptions \cite{ref-clip}. We use the pretrained image representation as one input to the hand forecast.

Robotics studies further examine how pretrained visual features transfer
to control. R3M learns a representation from human video for downstream
manipulation \cite{ref-r3m}; VIP derives visual representations and rewards
through value-based pretraining \cite{ref-vip}. VC-1 and CortexBench evaluate
representations across diverse embodied tasks \cite{ref-vc1}, while Voltron
combines language-conditioned visual reconstruction with visually
grounded language prediction \cite{ref-voltron}. We examine both geometric forecast accuracy on EgoDex and closed-loop task success in simulation.

DINOv2 learns transferable visual features through self-supervision
\cite{ref-dinov2}. Register tokens address artifacts in visual transformer
feature maps \cite{ref-registers}. We use the released register-equipped Base
encoder without updating its parameters. All matched runs reuse the same cached visual features.

\subsection{World-action prediction and visual
reconstruction}\label{world-action-prediction-and-visual-reconstruction}

EgoWAM compares pixel, feature and motion targets for human--robot co-training \cite{ref-egowam}. We hold the compact visual target fixed and vary the role of hand motion. In the human forecaster, two heads use the same observed context to predict future hand coordinates and a visual feature. Only the separate visual probe receives a future trajectory as input.

Representation autoencoders couple a representation encoder to an image decoder \cite{ref-rae}. Latent diffusion models \cite{ref-ldm} and diffusion transformers \cite{ref-dit} can generate images through a learned latent representation. Our visual objective operates directly on a pooled DINOv2 feature. We therefore evaluate it in feature space.

Auxiliary objectives also require a choice of relative loss scale.
Uncertainty-based weighting learns task weights \cite{ref-mtl_uncertainty}, while GradNorm
adapts them using gradient magnitudes \cite{ref-gradnorm}. We compare a zero visual-loss weight with a fixed positive weight under matched training conditions.

\begin{figure*}[!t]
\centering
\includegraphics[width=\linewidth,alt={EgoDex forecasting model. The action head predicts hand coordinates at all 60 query times. The visual head uses the hidden state at the final time query to predict the visual feature at two seconds. Solid arrows show the forward computation and residual connections. Dashed orange arrows indicate gradients to trainable parameters. The encoder is frozen. The photographs show recorded observations; hand symbols illustrate the output geometry.}]{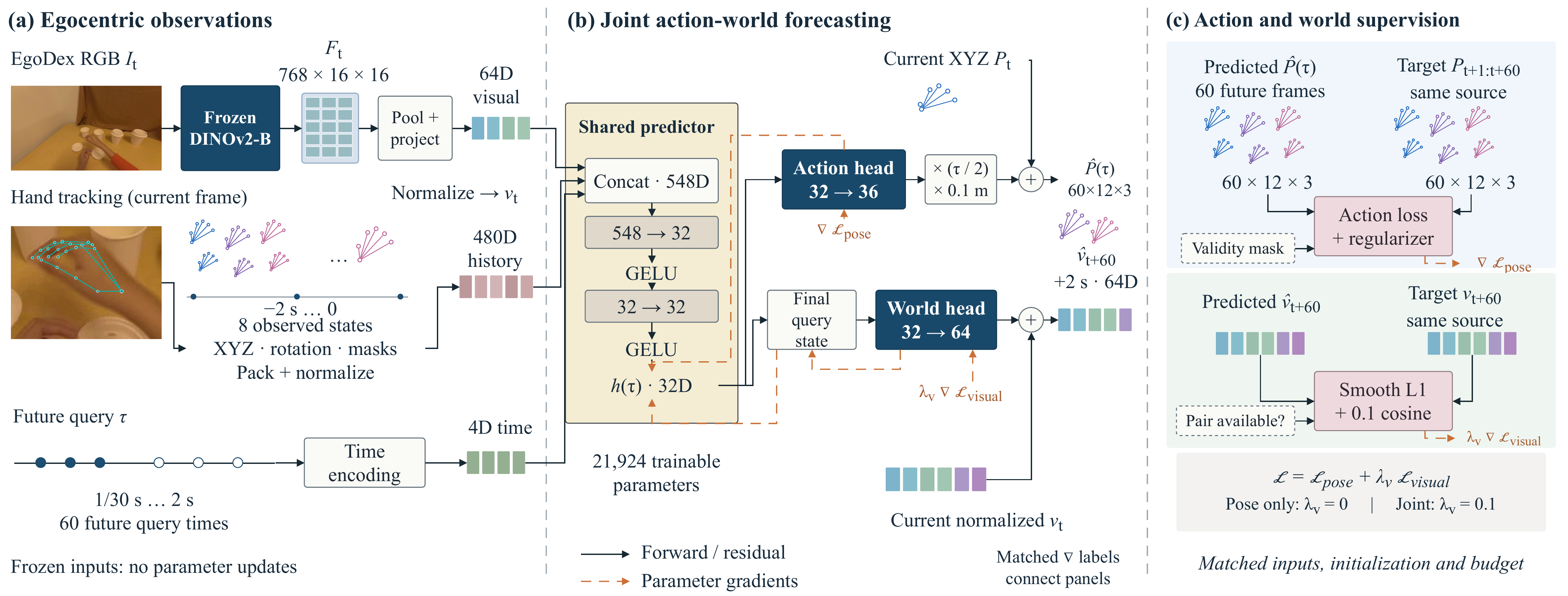}
\caption{EgoDex forecasting model. The action head predicts hand coordinates at all 60 query times. The visual head uses the hidden state at the final time query to predict the visual feature at two seconds. Solid arrows show the forward computation and residual connections. Dashed orange arrows indicate gradients to trainable parameters. The encoder is frozen. The photographs show recorded observations; hand symbols illustrate the output geometry.}
\label{fig:architecture}
\end{figure*}

\section{Task and data}\label{task-and-data}

\subsection{Observation and forecast}\label{observation-and-forecast}

At anchor frame \(t\), the predictor receives one RGB image and eight hand states. The offsets of the hand observations, in frames relative to \(t\), are
\[
(-60,-51,-43,-34,-26,-17,-9,0).
\]
At 30~Hz, this history spans two seconds. Each hand state contains 36 position coordinates for the two wrists and ten fingertips, 12 values encoding the two wrist rotations, and 12 landmark validity bits. Stacking the eight states gives a 480-dimensional history vector.

Let \(k=1,\ldots,60\) index future frames and \(j=1,\ldots,12\) index landmarks. The elapsed time at frame \(t+k\) is \(\tau_k=k/30\), expressed numerically in seconds. Write \(P_t\in\mathbb{R}^{12\times3}\) for the observed positions at the anchor and \(P^*_{t+k}\) for the corresponding recorded future positions. The forecast \(\hat P(\tau_k)\) has the same shape. Stacking all 60 forecasts gives a \(60\times12\times3\) trajectory. Hand rotations enter only the observed history.

We express all positions in the camera coordinate system fixed at the anchor. Evaluation uses these coordinates directly, without per-frame root or Procrustes alignment. The predictor uses observations through frame \(t\). Later images can supply training targets. The inputs contain neither task text nor object annotations.

\subsection{Coverage and
compatibility}\label{coverage-and-compatibility}

Throughout this paper, \textbf{full-data training} means using every
eligible training window in the fixed available EgoDex inventory after
session separation and geometric validity checks. Of 777,244
candidate windows, 762,777 remain after preserving the held-out
sessions; 11,712 of these fail the validity criteria.

The full training set includes 689,806 confidence-filtered windows and 61,259 additional windows admitted by the compatibility rule, for a total of 751,065.

The compatibility rule applies only to sources missing all 12 required
confidence fields. For these sources, geometric validity is determined
from finite, valid rigid transforms and valid parent wrists, without
applying a confidence threshold. Missing confidence is never imputed as
1. Supplied confidence values must be finite and within \([0.5,1]\);
partially missing fields remain masked. Existing camera, temporal and
anchor-validity checks are preserved.

\subsection{Splits and visual pairing}\label{splits-and-visual-pairing}

Training and held-out samples are separated by the saved session
inventory. Subject identities were not used to define splits. We select checkpoints by the 12-landmark distance on a fixed internal set of 96 windows. The main development set contains 36,596 windows from 68 tasks. It
has been inspected during experimentation and serves as the development evaluation. The official test set was not used.

A visual target is paired only when a same-source cached feature exists
at exactly \(t+60\). The 570,864 pairs cover 76.01\% of training
windows. The other 180,201 windows still train pose prediction; their
visual loss is masked. Future targets do not change the observed input.

\section{Joint world-action
formulation}\label{joint-world-action-formulation}

Figure~\ref{fig:architecture} connects recorded EgoDex observations to an action head and a
world-feature head through a shared predictor. In the current
implementation, action means a sequence of human wrist and fingertip
positions. The visual target is a normalized 64-dimensional feature two seconds after the anchor.
This is an observation-conditioned joint forecast. The predicted action
is not fed into the world head, and the 64-dimensional output does not encode
explicit object dynamics or support RGB decoding.

\subsection{Egocentric visual interface and shared
state}\label{egocentric-visual-interface-and-shared-state}

The DINO branch uses frozen DINOv2 Base weights with registers through
the RAE encoder interface. This interface removes the final LayerNorm
affine parameters and applies the released RAE latent statistics before
pooling. In the evaluated
cache pipeline, a 128-pixel preprocessing stage precedes 224-pixel
encoder input. The spatial feature grid \(F_t\in\mathbb{R}^{768\times16\times16}\) is
reduced by fixed \(3\times5\) pooling and a fixed seeded projection to
64 dimensions. Let \(f_t\) denote this 64-dimensional projected feature and \(v_t=(f_t-\mu_f)/\sigma_f\) its normalized value. Subtraction and division act componentwise using training-set statistics. Let \(\bar H_t\) denote the observed hand history normalized using its own training statistics. For each query time \(\tau\in\{\tau_1,\ldots,\tau_{60}\}\), the predictor concatenates the history, image feature and time encoding:
\begin{equation}
\begin{aligned}
x(\tau)&=[\bar H_t,\,v_t,\,q(\tau)]\in\mathbb{R}^{548},\\
q(\tau)&=[\tau,\tau^2,\sin(\pi\tau),\cos(\pi\tau)].
\end{aligned}
\end{equation}
The three inputs contribute 480, 64 and four values, respectively. Two linear layers of widths 32 and 32, each followed by GELU, produce the shared state \(h(\tau)\). All 60 queries use the same observed context and can be evaluated in parallel.

\subsection{Action and world-feature
outputs}\label{action-and-world-feature-outputs}

The pose head maps the shared state to a dimensionless residual \(D(\tau)\). With the position scale \(s_p=0.1\,\mathrm{m}\), the forecast is
\begin{equation}
\begin{aligned}
D(\tau)&=\operatorname{reshape}_{12\times3}(W_p h(\tau)+b_p),\\
\hat P(\tau)&=P_t+s_p\,(\tau/\tau_{60})\,D(\tau).
\end{aligned}
\end{equation}
Here \(W_p\) and \(b_p\) are the weights and bias of the 36-output pose head. The ratio \(\tau/\tau_{60}\) reaches one at the final query. The 64-output visual head uses only that final hidden state, with weights \(W_v\) and bias \(b_v\):
\begin{equation}
\hat v_{t+60}=v_t+W_v h(\tau_{60})+b_v.
\end{equation}

The two shared layers and both output heads contain 21,924 parameters in total.
With zero-initialized output layers, the initial predictions repeat the current pose and current feature. Both training conditions use the same architecture.
``Action-only'' denotes pose supervision; ``Joint'' denotes action plus
world-feature supervision. Both models receive the same current image feature. Only the joint model trains the visual head.

\begin{table*}[!t]
\centering
\caption{Main comparison on the same EgoDex development set. Endpoint and trajectory errors (top) and wrist displacements (bottom) are in millimeters. Smaller errors indicate more accurate forecasts. Learned models use the full
eligible training set. Model errors give the mean and sample standard deviation (SD) across four paired seeds. Forecast displacements give the mean. Both learned models use the same inputs, initialization and
training budget.}
\label{tab:result1}
\footnotesize
\setlength{\tabcolsep}{3pt}
\begin{tabular}{@{}lrrr@{}}

\toprule\noalign{}
Method & Wrist ADE & Wrist FDE & MPJPE-12 \\
\midrule\noalign{}

Stationary wrist and fingertips & 40.685 & 60.262 & 48.056 \\
Constant velocity & 90.566 & 170.297 & 113.126 \\
Action-only & \(38.638\pm 0.056\) & \(56.480\pm 0.197\) & \(45.413\pm 0.035\) \\
Joint & \(38.631\pm 0.057\) & \(56.460\pm 0.189\) & \(45.413\pm 0.041\) \\
\midrule
\multicolumn{4}{@{}l}{\emph{Two-second wrist displacement, grouped by recorded distance}}\\
Wrist group & Recorded & \multicolumn{2}{c@{}}{Action-only forecast} \\
All eligible wrists & 60.26 & \multicolumn{2}{c@{}}{13.07} \\
Recorded displacement \(\geq80\) mm & 147.23 & \multicolumn{2}{c@{}}{22.12} \\
\bottomrule
\end{tabular}
\end{table*}

\subsection{Masked losses and matched
training}\label{masked-losses-and-matched-training}

For one window, let \(m_{kj}\in\{0,1\}\) indicate whether landmark \(j\) is valid at query \(k\), and let \(N=\sum_{k,j}m_{kj}>0\). Write \(\hat p_{kj}\) and \(p^*_{kj}\) for row \(j\) of \(\hat P(\tau_k)\) and \(P^*_{t+k}\), respectively, and \(p_{0j}\) for row \(j\) of the observed \(P_t\). The three-vectors are expressed in meters. With smoothing parameter \(\epsilon=0.001\,\mathrm{m}\), the window loss is
\begin{equation}
\begin{aligned}
e_{kj}&=\hat p_{kj}-p^*_{kj},\qquad
r_{kj}=(\hat p_{kj}-p_{0j})/s_p,\\
\rho(e)&=\bigl(\sqrt{\lVert e\rVert_2^2+\epsilon^2}-\epsilon\bigr)/s_p,\\
\ell_p&=\frac{1}{N}\sum_{k,j}m_{kj}
\left[\rho(e_{kj})+\tfrac13\lVert r_{kj}\rVert_2^2\right].
\end{aligned}
\end{equation}
The first term measures position error. The second penalizes displacement from the observed landmark, averaging the squared residual over the three coordinates. Averaging \(\ell_p\) over windows in the batch gives \(\mathcal L_{\mathrm{pose}}\).

For each paired window, the visual loss is the mean smooth L1 loss across the 64 feature coordinates plus 0.1 times the cosine distance between the predicted and target normalized features. We average this loss over paired windows in the batch to obtain \(\mathcal L_{\mathrm{visual}}\), with zero contribution if the batch has no pairs. The training objective is
\begin{equation}
\mathcal L=\mathcal L_{\mathrm{pose}}+\lambda_v\mathcal L_{\mathrm{visual}},
\end{equation}
where \(\lambda_v=0\) for action-only training and \(\lambda_v=0.1\) for joint training.

Both conditions use the same windows, initialization, batches and checkpoint criterion for each of four paired seeds. We train for two epochs (11,736 updates per model) with AdamW \cite{ref-adamw}, a batch size of 128, a learning rate of \(3\times10^{-4}\), weight decay of \(10^{-4}\) and gradient clipping at 1.

The pose loss updates the action head and shared layers at every valid query time. The visual loss updates the visual head and the same shared layers at the final query. A single endpoint feature can therefore influence the trajectory through shared parameters, although different hand paths can lead to similar endpoint images.

\section{Forecasting hand trajectories on EgoDex}\label{evaluation-and-results}

Table~\ref{tab:result1} compares two learned forecasts with stationary and constant-velocity baselines on the same development windows.

\subsection{Metrics and aggregation}\label{metrics-and-aggregation}

For each valid landmark and frame, the Euclidean error in millimeters is \(d_{kj}=1000\lVert e_{kj}\rVert_2\) when \(e_{kj}\) is expressed in meters. Wrist average displacement error (ADE) averages \(d_{kj}\) over the two wrists and all valid future frames. Wrist final displacement error (FDE) uses the wrists at \(k=60\). Mean per-joint position error (MPJPE-12) averages \(d_{kj}\) over the 12 wrist and fingertip landmarks and all valid future frames.

Metrics are averaged within windows, then hierarchically across
episodes, sessions and tasks with equal weight at each level. Missing
entries are excluded, not set to zero. Endpoint metrics never substitute
the last valid earlier frame. The combined cohort has 36,596 valid
windows for trajectory errors and 36,469 for exact wrist endpoints. Predictions are deterministic (\(K=1\)); no
best-of-\(K\) selection is used. The sample SD across four seeds
measures training variation, not a confidence interval over independent
sessions. For selected paired contrasts we also report 95\% Bayesian bootstrap intervals over sessions. Each of 10,000 draws assigns one independent exponential weight of mean one to each session, shared across conditions and tasks. We renormalize session weights within each task and average the task estimates equally. Intervals use the 2.5th and 97.5th percentiles with tasks, folds and fitted models fixed.

\subsection{Matched full-data
comparison}\label{matched-full-data-comparison}

The stationary baseline repeats the observed anchor position. Constant velocity
extrapolates each landmark using the velocity between its most recent
valid historical observation and the anchor; it uses zero velocity when
no valid pair is available. The two learned models use identical data
and architecture. Action-only uses the pose loss; Joint adds the
future-feature loss. Both models receive the current image and hand
history.

The action-only model improves all three trajectory metrics over the stationary baseline (Table~\ref{tab:result1}). At two seconds, wrist endpoint error falls by 3.78~mm. The joint model improves on stationary prediction by 3.80~mm (session interval 3.39 to 4.21~mm). Extrapolating the last observed velocity gives the largest endpoint error. Adding the future-feature loss changes wrist endpoint error by just 0.02~mm at the tested loss weight.

For the action-only model, training on the expanded inventory lowered MPJPE-12 on development windows without confidence fields (73.74 to 71.29~mm) and slightly raised it on confidence-filtered windows (40.96 to 41.09~mm). Both training runs covered two epochs with normalization fitted to their respective training data.

\begin{figure*}[!t]
\centering
\includegraphics[width=\linewidth,alt={Three rows of six recorded frames each. The recorded and forecast wrists stay close during the small movement in A and separate during the larger movements in B and C.}]{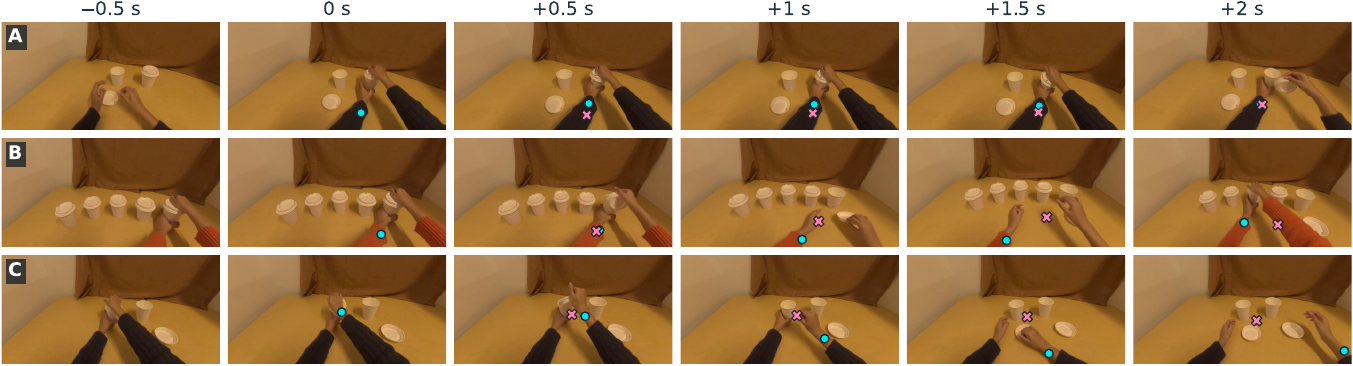}
\caption{Wrist forecasts from the four-task probe in three held-out EgoDex lid episodes \cite{ref-egodex}. Cyan dots mark recorded positions and pink crosses mark forecasts made at 0~s. Rows A and B show the left wrist, and row C shows the right wrist.}
\label{fig:lid-reach}
\end{figure*}

\subsection{Predicting the future visual feature}
All four pose-selected checkpoints were evaluated on the same 28,336 development windows with a paired two-second target (63 tasks). We averaged errors within episodes, then across sessions and tasks as in the trajectory evaluation. With the model, image and target fixed, we replaced each eight-state hand history with one from another held-out session. Each of 20 permutations used every history once and was shared across checkpoints.

\begin{table}[!ht]
\centering
\caption{Two-second DINO feature prediction on matched EgoDex windows. Model errors are means across four training seeds. Lower is better.}
\label{tab:visual-head}
\footnotesize
\begin{tabular}{@{}lrr@{}}
\toprule
Forecast & Smooth L1 & Cosine distance \\
\midrule
Copy current feature & 0.1850 & 0.2045 \\
Joint, aligned history & 0.1825 & 0.2029 \\
Joint, shuffled history & 0.1884 & 0.2084 \\
\bottomrule
\end{tabular}
\end{table}

Across seeds, aligned histories reduced smooth L1 error by 1.32\% (session interval 1.17\% to 1.46\%) relative to copying the current feature, while shuffled histories increased it by 1.85\% (Table~\ref{tab:visual-head}). The ordering held for every seed and also when histories were exchanged only between episodes of the same task. After averaging across seeds within each task, aligned histories beat copying on 49 of 63 tasks. Exchanging a history alters its motion, anchor pose and tracking validity together.

\subsection{Wrist displacement across the forecast horizon}

Table~\ref{tab:result1} also reports the three-dimensional distance from the anchor to the wrist endpoint. The action-only model predicts only a small fraction of the displacement during long reaches. For wrists moving less than 20~mm, the mean forecast is closer to the recorded distance (4.91 versus 6.01~mm). These groups use the realized displacement and are defined after prediction.

\section{Can a hand trajectory inform a visual forecast?}

The main experiment varies the supervision used to train the action predictor. Here we ask a different question. For the same observed context, how much does knowing the subsequent hand motion improve a visual forecast? We compare recorded and predicted trajectories as inputs to a separate regression model.

\subsection{Inputs and separation of sessions}

We use 927 eligible windows from 183 recording sessions covering four tasks: adding or removing a lid, assembling or disassembling blocks, cleaning a surface, and coloring. These windows came from the training partition of the smaller cached experiment. Five outer folds hold out complete sessions. Each window is scored on its outer held-out fold. Forecast-conditioned predictions are evaluated for each pose seed. The target is the change in a 64-dimensional pooled full-image feature from the anchor to the exact two-second endpoint. A zero change corresponds to copying the current feature.

The context includes the current image features and the observed hand history with validity indicators. The trajectory condition adds the displacement of each of the 12 landmarks relative to the anchor at four horizons: 0.5, 1.0, 1.5 and 2.0~s. We fit ridge regressions with a fixed unit penalty. Feature scaling and target normalization use only the training sessions of the corresponding outer fold. Window-level mean squared errors are averaged within sessions and then across sessions and tasks. The resulting scores measure squared error in normalized feature space.

\subsection{Recorded and forecast motion}

Figure~\ref{fig:lid-reach} shows three examples selected from the four-task probe. In A, the forecast closely matches the magnitude of the two-second displacement (40~mm predicted, 41~mm recorded). In B and C, it predicts about 68~mm for reaches of 221 and 331~mm, respectively.

The recorded condition supplies future landmark positions and validity masks. For the forecast condition, four inner folds generate out-of-fold trajectories to fit the visual regression. For each outer fold we fit a separate compact action predictor using the four-task training sessions. Its held-out trajectories use only anchor-time inputs. Table~\ref{tab:result1} reports a separate full-data model. We average errors over four pose seeds. Shuffling forecasts within task and partition tests whether matching a trajectory to its own window matters.

\begin{table}[!ht]
\centering
\caption{Predicting visual feature change at two seconds. Normalized mean squared error is macro-averaged over four tasks or reported for the lid task alone. Lower values indicate smaller errors. Recorded trajectories include future validity masks.}
\label{tab:visual-probe}
\footnotesize
\setlength{\tabcolsep}{3pt}
\begin{tabular}{@{}lrr@{}}
\toprule
Probe input & Four tasks & Lid task \\
\midrule
Copy current feature & 0.436436 & 0.690897 \\
Observed context & 0.389063 & 0.579985 \\
Context + recorded future mask & 0.389087 & 0.577395 \\
Context + shuffled forecast & 0.389552 & 0.581673 \\
Context + forecast & 0.389050 & 0.580010 \\
Context + future mask and displacements & 0.373793 & 0.537339 \\
\bottomrule
\end{tabular}
\end{table}

Adding future tracking-validity masks to observed context changed the four-task MSE by less than 0.01\% (Table~\ref{tab:visual-probe}). With the same masks included, adding recorded displacements reduced error by 3.93\% (session interval 3.12\% to 4.72\%). The full input reduced error by 3.92\% relative to context alone. None of the 927 windows had valid labels for all 12 landmarks at all four horizons, so the comparison uses all windows with invalid displacements masked. The probe's forecast trajectories lowered the four-task error by only 0.0034\% relative to context alone and slightly increased it on lids.

\section{Does human hand history help a robot policy?}
\label{sec:robottest}

We evaluate policies on two simulated RoboTwin~2.0 tasks: grasping two bottles and handing over a block between arms \cite{ref-robotwin}. We measure full-task success on 100 held-out scenes per task and training seed. The bottle comparisons share the same 48 ALOHA demonstrations and evaluation scenes. One spatial policy also uses demonstrations from three additional robots. We inspected the bottle suite during policy design.

\begin{figure}[!t]
\centering
\begin{tikzpicture}[x=1cm,y=1cm, font=\footnotesize, box/.style={draw=black!45,rounded corners=2pt,align=center,inner sep=4pt}, flow/.style={->,thick}]
\node[box,fill=blue!6,text width=3.1cm] (robot) at (-1.95,2.2) {Robot RGB and arm states\\Robot input adapter};
\node[box,fill=green!8,text width=3.1cm] (human) at (1.95,2.2) {Human RGB and hand history\\Human input adapter};
\node[box,fill=black!5,text width=4.6cm] (shared) at (0,1.1) {Shared temporal Transformer};
\node[box,fill=blue!6,text width=3.1cm] (action) at (-1.95,0) {Robot action head\\16 bimanual command steps};
\node[box,fill=green!8,text width=3.1cm] (world) at (1.95,0) {Future-feature head\\Predicted 64-D DINO feature};
\draw[flow] (robot.south) -- (robot.south |- shared.north);
\draw[flow,dashed] (human.south) -- (human.south |- shared.north);
\draw[flow] (action.north |- shared.south) -- (action.north);
\draw[flow,dashed] (world.north |- shared.south) -- (world.north);
\end{tikzpicture}
\caption{Robot transfer experiment. The solid path maps robot observations to executable commands. The dashed paths add human observations and a future-feature loss during training. The EgoDex trajectory forecast in Fig.~\ref{fig:architecture} is evaluated separately.}
\label{fig:robot-transfer}
\end{figure}
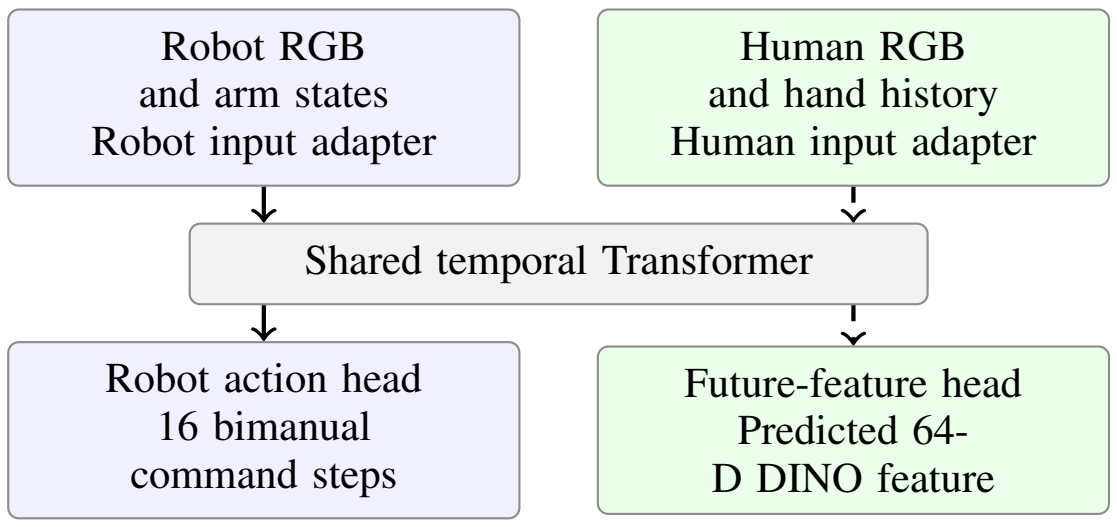

\subsection{Training and execution}
Figure~\ref{fig:robot-transfer} shows the connection between embodiments. A human example contains eight observed hand states and the current image feature. Its target is the DINO feature two seconds later in the same recording. A robot example contains eight robot observations. Its visual target is 16 demonstration steps ahead. During human--robot training, the robot action loss and both domains' feature losses update the shared Transformer. At deployment, the action head predicts 16 steps of end-effector and gripper commands. The robot executes the first eight steps before the policy receives a new observation.

\subsection{Hand-history supervision and task success}
The robot-only policy learns robot actions and future visual features. For the comparison with human supervision, we replace an auxiliary robot feature batch with EgoDex examples. We either retain each hand history with its video or shuffle the histories while keeping the current and future human images paired. All three policies see the same robot action batches within each task and seed.

\begin{table}[!ht]
\centering
\caption{Closed-loop success (\%) on the same 100 held-out scenes per task. Entries are means and sample SDs across three training seeds. Aligned and shuffled refer to the hand history paired with each human video during training.}
\label{tab:hand-history}
\footnotesize
\begin{tabular}{@{}lrr@{}}
\toprule
Supervision & Bottle grasping & Block handover \\
\midrule
Robot only & \(16.0\pm2.0\) & \(39.0\pm8.7\) \\
Robot + aligned history & \(15.0\pm2.6\) & \(37.7\pm4.0\) \\
Robot + shuffled history & \(14.7\pm1.5\) & \(41.7\pm9.0\) \\
\bottomrule
\end{tabular}
\end{table}

Aligned human histories gave no consistent gain over robot-only supervision (Table~\ref{tab:hand-history}). In seed order (20261110, 20261111, 20261112), aligned minus shuffled success rates differed by \(+3,+2,-4\) percentage points on bottle grasping and \(-18,0,+6\) on handover.

\subsection{Bottle policy controls}
An initial single-seed bottle experiment tested the pairing of future \emph{visual targets}, with hand histories kept aligned to the current image. Four policies started from the same robot history checkpoint and received 5,000 updates. Action supervision alone gave 14 successes in 100 scenes. Adding robot future-feature supervision gave 18. Replacing the auxiliary robot batch with samples from 8,138 EgoDex training pairs gave 16 with paired targets and 14 with shuffled targets. The robot-only visual and paired-human results for this seed also contribute to Table~\ref{tab:hand-history}.

A separate robot-only comparison examined visual resolution and training data. Keeping a \(3\times5\) spatial grid gave 25 successes in the same 100 scenes, compared with 21 for the width-matched compact policy. Adding 50 demonstrations each from ARX-X5, Piper and UR5 to 48 ALOHA demonstrations gave 32 successes with the spatial policy on ALOHA. The pooled policy gained 16 scenes and lost nine relative to ALOHA-only training (two-sided exact McNemar \(p=0.230\)). This single-seed comparison varies both demonstration count and embodiment diversity.

\section{Discussion}\label{discussion}

\subsection{What the forecasts capture}
The displacement penalty in the pose loss favors short forecasts. Consider one target at distance \(D\), with the direction known. Ignoring the 1~mm smoothing term, predicting distance \(d\) costs \(\lvert D-d\rvert/s_p+d^2/(3s_p^2)\), minimized at \(d=\min(D,1.5s_p)\). With \(s_p=0.1\)~m this idealized optimum saturates at 150~mm. In the four-task probe, removing the penalty increased mean predicted landmark displacement from 11.28 to 18.72~mm over four horizons, still well below the recorded 69.62~mm. This ablation concerns the compact predictor. The contribution of the penalty to underreach in the full-data model remains unmeasured.

The probe provides a complementary result. Recorded displacements add predictive information beyond the observed context and future tracking masks. The target covers the entire frame. Hand appearance, viewpoint and object state can all change it; the lid experiment measures no separate object-state outcome. A deterministic trajectory forecast can re-express the observations for the linear probe, but it supplies no independent observation of the future. Its negligible incremental gain therefore measures its value to this probe. It does not by itself establish that trajectory error caused the gap.

\subsection{Scope of the robot results}
Human hand history enters the robot policy through an auxiliary feature-prediction objective. Across the two simulated tasks, preserving its alignment with the video gave no consistent improvement over shuffling it. All policies within a task share the initial robot core: a trained bottle checkpoint for grasping and a common random initialization for handover. Comparisons therefore pair policies within each task. Success varied most across seeds on handover.

The hand-history comparison uses three training seeds per task. The bottle spatial-feature and cross-embodiment comparisons each use a single training seed. For EgoDex, the four paired seeds measure training variation on fixed development sessions. Geometric validity extends coverage to recordings without confidence fields, whose tracking quality remains unmeasured.

\section{Conclusion}\label{conclusion}

EgoHandWAM reduced wrist endpoint error on EgoDex but underestimated long reaches. The visual head forecast two-second features more accurately than copying the current feature when the hand history matched the image. In a separate probe, recorded displacements improved the whole-image feature forecast after accounting for future tracking validity. Trajectories from the probe's compact action predictor added little. On simulated bottle grasping and block handover, training with aligned human histories gave no consistent increase in robot task success.

\end{document}

%% file: header.tex
\renewcommand{\scshape}{\upshape}
\makeatletter
\renewcommand{\section}{\@startsection{section}{1}{\z@}{3ex plus 1ex minus 1ex}{0.7ex plus .5ex}{\normalfont\normalsize\bfseries\centering}}

\g@addto@macro\UrlBreaks{\do\a\do\b\do\c\do\d\do\e\do\f\do\g\do\h\do\i\do\j\do\k\do\l\do\m\do\n\do\o\do\p\do\q\do\r\do\s\do\t\do\u\do\v\do\w\do\x\do\y\do\z\do\0\do\1\do\2\do\3\do\4\do\5\do\6\do\7\do\8\do\9}
\makeatother

%% file: authors.tex
\author{%
{\small\textbf{Kailin Wang}$^{1}$ \quad
\textbf{Haoxiang Jie}$^{1}$ \quad
\textbf{Yaoyuan Yan}$^{1}$ \quad
\textbf{Zhiyou Heng}$^{1}$ \quad
\textbf{Zhaosong Li}$^{2,3}$}\\[0.35em]
{\small $^{1}$AI Lab, Country Garden Services Group}\\
{\small $^{2}$School of Mechanical Science and Engineering, Huazhong University of Science and Technology}\\
{\small $^{3}$Omni AI}\\[0.35em]
{\small \href{mailto:wangkailin417@gmail.com}{\texttt{wangkailin417@gmail.com}}\quad
\href{mailto:jiehaoxiang@bgyfw.com}{\texttt{jiehaoxiang@bgyfw.com}}}
}